\documentclass[runningheads]{llncs}
\usepackage[T1]{fontenc}
\usepackage{graphicx,verbatim}
\PassOptionsToPackage{table}{xcolor} 
\usepackage{amssymb,amsmath}
\usepackage[labelfont=bf]{caption}
\usepackage{subcaption}
\usepackage{xcolor}
\usepackage{multirow}
\usepackage{multicol}
\usepackage{booktabs}
\usepackage{cite}
\usepackage[breaklinks]{hyperref} 
\usepackage{mathrsfs}
\usepackage{listings}
\usepackage{parskip}
\usepackage[shortlabels]{enumitem}
\usepackage{tcolorbox}
\usepackage{dsfont}
\usepackage{algorithm}
\usepackage{algpseudocode}
\usepackage{fancyvrb}
\usepackage{array}
\usepackage{makecell}
\usepackage{pifont}
\newcommand{\cmark}{\ding{51}}%
\newcommand{\xmark}{\ding{55}}%

\graphicspath{{images/}}
\usepackage{color}

\begin{document}
\title{Specialised or Generic? Tokenization Choices for Radiology Language Models}

\titlerunning{Specialised or Generic Tokenization}

\author{Hermione Warr\inst{1}, 
Wentian Xu\inst{1},
Harry Anthony\inst{1},
Yasin Ibrahim\inst{1}, 
Daniel R. McGowan\inst{2,3}, 
Konstantinos Kamnitsas\inst{1,4,5}}
\authorrunning{H. Warr et al.}

\institute{Department of Engineering Science, University of Oxford, Oxford, UK \\
\email{\{first\_name.last\_name\}@eng.ox.ac.uk}
\and
Department of Oncology, University of Oxford, UK
\and
Department of Medical Physics and Clinical Engineering, Oxford University Hospitals NHS FT, Oxford, UK
\and
Department of Computing, Imperial College London, London, UK
\and
School of Computer Science, University of Birmingham, Birmingham, UK
}

\maketitle              
\begin{abstract}

The vocabulary used by language models (LM) - defined by the tokenizer - plays a key role in text generation quality. 
However, its impact remains under-explored in radiology.
In this work, we address this gap by systematically comparing general, medical, and domain-specific tokenizers on the task of radiology report summarisation across three imaging modalities.
We also investigate scenarios with and without LM pre-training on PubMed abstracts. Our findings demonstrate that medical and domain-specific vocabularies outperformed widely used natural language alternatives when models are trained from scratch.
Pre-training partially mitigates performance differences between tokenizers, whilst the domain-specific tokenizers achieve the most favourable results.
Domain-specific tokenizers also reduce memory requirements due to smaller vocabularies and shorter sequences.
These results demonstrate that adapting the vocabulary of LMs to the clinical domain provides practical benefits, including improved performance and reduced computational demands, making such models more accessible and effective for both research and real-world healthcare settings.
Code available at: \href{https://github.com/hermionewarr/Scribe}{GitHub}.


\end{abstract}
\section{Introduction}
\label{sec:intro}
Radiologists are facing mounting challenges managing the growing volume of imaging data. 
The integration of artificial intelligence (AI) for assisting radiology report generation has gained attention as a potential solution. 
The advent of Transformers \cite{vaswani_attention_2017} significantly advanced language models (LMs), prompting interest in their application to radiology reporting~\cite{Liu_review_2023}.
As LMs become increasingly accessible and capable, adapting them for clinical use remains non-trivial. A key challenge lies in ensuring factual accuracy in generating complex radiological text under computational constraints.
Strategies to address this include adapting model architectures \cite{meshed-memory_T_2020} and biomedical domain pre-training \cite{Gu_2022}. An important, yet often overlooked component, is the \emph{tokenizer}, which defines the \emph{vocabulary} the model operates on. Tokenization determines how text is represented in an LM, influencing both model performance and computational efficiency.

In prior work, most radiology LMs were built on general-purpose tokenizers trained on natural language corpora, such as those from GPT-2 or LLaMA2 \cite{Liu_review_2023}. Others use generic biomedical vocabularies like PubMedBERT, trained on PubMed abstracts \cite{Gu_2022}, and some models were developed with a specific medical dataset vocabulary \cite{Liu_review_2023, Boecking_2022_ms}. 
Clinical text diverges from general English in syntax, terminology, and structure. It includes specialised vocabulary, abbreviations, and structured content like diagnostic codes and measurements \cite{singh_2024}. This can challenge general tokenizers and limit downstream performance.
Moreover, most studies focused on chest X-ray reports, while other modalities remain under-explored \cite{Hamamci_ctclip_2024, Liu_review_2023}. 
This issue is compounded by the computational demands of large LMs, limiting research in resource-constrained settings, especially on 3D scans like PET-CTs.
This motivates research into smaller, task-specific models that require less memory and compute compared to large-scale general purpose models.
The choice of tokenizer - and by extension, vocabulary - affects all of the above factors. It determines how efficiently medical language is represented, impacting both model performance and computational requirements.

In this study we develop a range of domain-specific and biomedical vocabularies, and compare them against general natural-language vocabularies for the task of radiology report summarisation. We also examine scenarios with and without LM pre-training. 
We assess performance and memory requirements across 3 datasets that span 3 imaging modalities: X-rays using MIMIC-CXR\cite{mimic_paper}, CTs with CT-RATE\cite{Hamamci_ctclip_2024}, and a private oncology PET-CT database.
Our findings demonstrate that data-specific tokenization improves both performance and memory efficiency across all datasets when LMs are trained from scratch. These insights inform best practices for developing radiology LMs, particularly in resource constrained settings.

\section{Methodology}
\label{sec:method}
\subsubsection{Problem Setting and Report Summarisation task:}

The task of generating a textual sequence $T$ can be framed as estimating the conditional distribution $p(T)$ as product of conditional probabilities, 
$
p(T) = \prod_{s=1}^{S} p(T_s\:|\:T_{<s};\:\theta).
$
Here $T$ is modelled as a sequence of word tokens $\{T_{1},...,T_{S}\}$, $T_{<s}$ is the set of tokens preceding $T_s$, and $\theta$ is the model parameters. 
A \emph{tokenizer} maps input text to a set of $S$ token vectors $\{T_{tok,s}\}_{1:S} \in \mathbb{R}^{S \times V}$, where $S$ is the sequence length and $V$ the size of the \emph{vocabulary}. 
Each $T_{tok,s}$ is embedded by a linear layer, so $T_{emb}=f_{T,emb}(T_{tok}) \in \mathbb{R}^{S \times D}$ is a set of S token embeddings with dimensionality $D$. 
The token embeddings are passed into a Transformer decoder with $N$ attention blocks, each with $H$ heads \cite{vaswani_attention_2017}. This outputs a probability distribution over the $V$ vocabulary tokens, indicating the most likely next token in the sequence.

\begin{figure}[t]
\centering
\includegraphics[width=\textwidth]{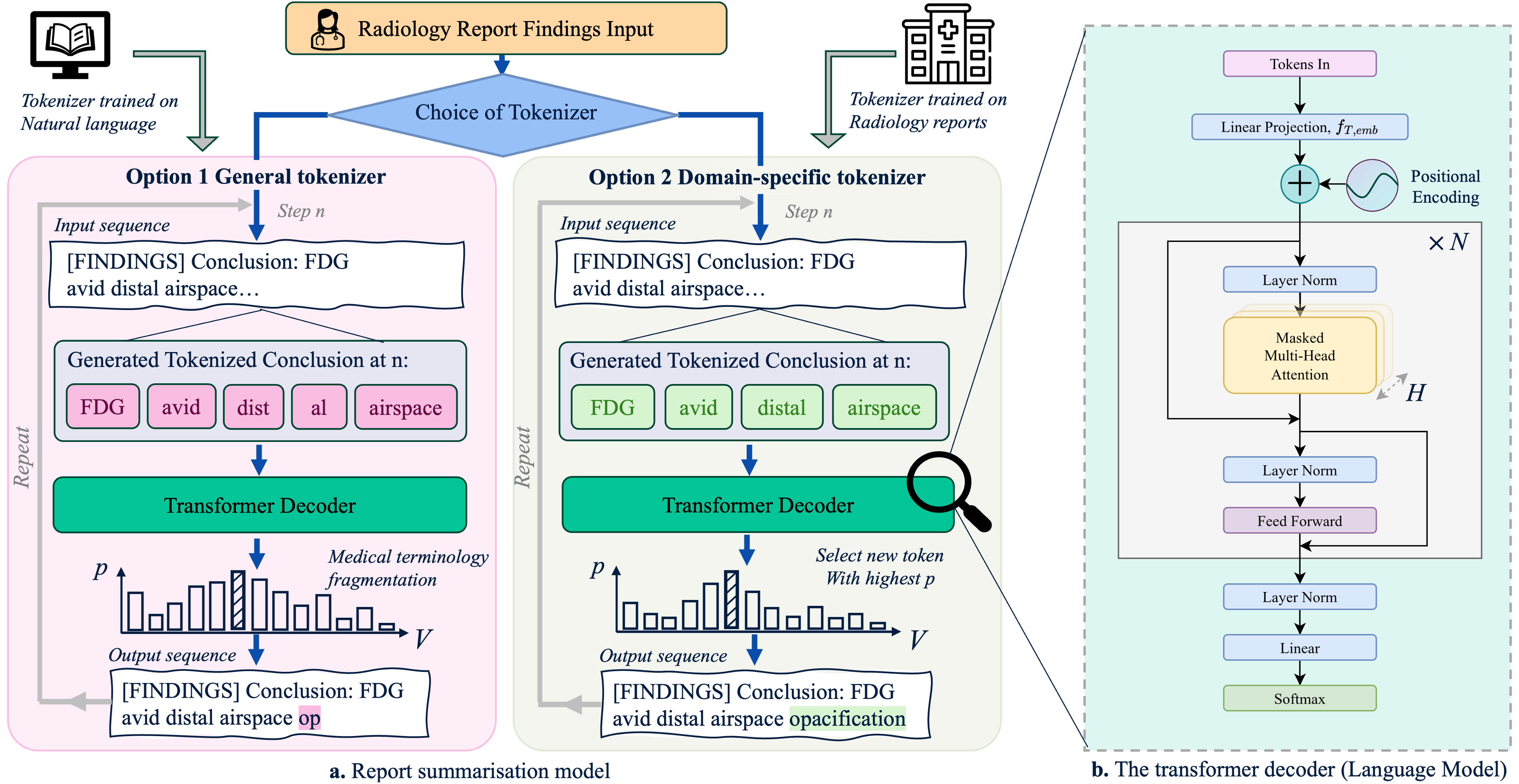}
\caption{
(a) An inference step for radiology report summarisation: predicting the \emph{Conclusion} from the \emph{Findings} section. The tokenized Findings and previously generated Conclusion tokens are passed to the autoregressive model (b), which outputs a vocabulary distribution to predict the next token. A general tokenizer requires more tokens and longer sequences, as out-of-vocabulary words (e.g., distal, opacification) are fragmented.
}
\label{fig:model}
\end{figure}

We evaluate the impact of the tokenizer on the task of \emph{radiology report summarisation} (Fig.~\ref{fig:model}). 
Specifically, we input the \emph{Findings} section of a radiology report to the model. We then train the LM to generate the \emph{Conclusion} of the report, which summarises the main findings.

\subsubsection{Tokenizers in depth:}

Tokenization is a preprocessing step in NLP that converts raw text into a sequence of smaller units called tokens, which serve as the atomic inputs to LMs.
Tokenizers define how text is broken down and map character sequences to a fixed \emph{vocabulary} - a set of unique tokens \emph{learned} from a training corpus \emph{prior} to LM training.
Without loss of generality, we focus on \emph{Byte Pair Encoding (BPE)} tokenizers~\cite{BPE_2016}, a widely-used tokenization algorithm.
Training a BPE tokenizer involves learning a series of merge operations from a text corpus. This process begins with a dictionary consisting of all individual characters present in the corpus. The most frequent adjacent pairs of symbols are iteratively merged to form new subword tokens, until either all words in the corpus are included or a predefined vocabulary size is reached \cite{BPE_2016}. The result is an ordered list of merge operations and a vocabulary, both of which are required at inference time to tokenize new inputs.
\
To \emph{apply} a trained BPE tokenizer, the input text is first split into individual characters. The tokenizer then applies the learned merge operations in order, combining frequent adjacent pairs into subwords until no further merges are possible.
\
The vocabulary learned by the tokenizer depends on the text the tokenizer is trained on. We define \emph{domain-specific tokenizers} as those trained exclusively on text from a single domain, such as radiology reports, and \emph{general tokenizers} as those trained on broad, multi-domain corpora. Given a maximum size of allowed vocabulary, training a tokenizer with larger, more varied text, results in vocabularies where rare words are not represented, and instead broken down into smaller subword tokens (fragmentation). 
This trade-off is illustrated in Figure~\ref{fig:model}a, which highlights how the same input is tokenized under different vocabularies. 
Once trained, the tokenizer is applied to the training data, and the resulting tokens are used to train an LM.

\subsubsection{Vocabulary, sequence length, and required memory:}
\label{sec:method_mem}

Properties of the tokenizer affect the memory required by LMs. 
Studying this can inform the choice of appropriate tokenizer, or enable training with more limited hardware. 
To this end, we derive an estimation of GPU memory $M$ required during the training of a Transformer-based LM with the general architecture shown in Fig.~\ref{fig:model}. This estimate depends on 
the tokenizer's vocabulary size $V$, input sequence length $S$, batch size $B$, model hidden dimension $D$, number of heads $H$, and number of blocks $N$.
We can approximate the memory $M$ required by such an LM, in one training step, as a sum of four components: 
\begin{equation}
    M = M_{\text{act}} + M_{\theta} + M^{\theta}_{\text{grad}} + M_{\text{opt}}
\label{eq:mem_tot}
\end{equation}
where $M_{\text{act}}$ is memory for storing activations, 
$M_{\theta}$ for model parameters $\theta$, $M^{\theta}_{\text{grad}}$ for gradients of $\theta$, $M_{\text{opt}}$ for optimiser state. We focus on $M_{\text{act}}$ as it dominates other terms in common settings \cite{mem_paper}. We derive an approximation of $M_{act}$ as:
\begin{align}
    M_{\text{act}} &= 2BSV + 2BSD + N \Big[ 
        \underbrace{16BSD + 2(BS^2H)}_{attn. \ block} 
    \Big].
\label{eq:mem_act}
\end{align}
Detailed derivation of all terms is in Supplementary Sec.~1. 
The tokenizer determines vocabulary size $V$, which affects memory usage in the input and output layers of the LM (via the $2BSV$ term). Importantly, it also determines indirectly the LM's sequence length $S$, which impacts memory across all layers, including quadratic terms.
Words not in the tokenizer’s vocabulary are split into multiple tokens (Fig.~\ref{fig:model}a), increasing $S$ for a given input. As a result, general-purpose tokenizers alongside  larger $V$, often require $S$ than domain-specific ones, leading to higher overall memory demands.

\section{Experiments}
\subsection{Experimental Settings}
\label{sec:train}

\subsubsection{Data:}
Experiments were performed using two public datasets, MIMIC-CXR \cite{mimic_paper} and CT-RATE \cite{Hamamci_ctclip_2024}, along with a private dataset of radiology reports of cancer patients scanned with whole body PET-CT. 
Statistics about the datasets and the train, validation and testing splits are shown in Table~\ref{tab:dataset_stats}.
\
In all experiments, from each report we use the \emph{Findings} section, which details observations in the patient's scan, and the \emph{Conclusion} section, which summarises key findings.

\begin{table}[t]
    \centering
    \caption{Dataset statistics for the datasets used in this study.}
    \renewcommand{\arraystretch}{1.2} 
    \setlength{\tabcolsep}{5pt} 
    \resizebox{\textwidth}{!}{
    \begin{tabular}{lccc p{2.3cm}<{\centering} p{2.6cm}<{\centering} p{2cm}<{\centering} ccc}
        \toprule
        \textbf{Dataset} & \textbf{No. Reports} & \textbf{Patients} & \textbf{Studies}  
        & \multirow{2}{*}{\shortstack{\textbf{Findings Len.} \\ $\mu$ ($\sigma$)}} 
        & \multirow{2}{*}{\shortstack{\textbf{Conclusions Len.} \\ $\mu$ ($\sigma$)}} 
        & \multirow{2}{*}{\shortstack{\textbf{No. Unique} \\ \textbf{Words}}} 
        & \multicolumn{3}{c}{\textbf{Data split}} \\ 
        \cmidrule(lr){8-10}
        & & & & & & & \textbf{Train} & \textbf{Val} & \textbf{Test} \\
        \midrule
        MIMIC-CXR \cite{mimic_paper}  & 215,371 & 63,937 & 218,081 & 49 (23) & 25 (24) & 12,570 & 134,805 & 1078 & 1955  \\
        CT-RATE \cite{Hamamci_ctclip_2024}  & 22,999 & 21,304 & 25,692 & 196 (72) & 34 (29) & 9,783 & 22,587 & 1,526 & 1,564 \\
        PET-CT  & 45,717 & 27,323 & 45,781 & 147 (69) & 36 (25) & 14,663 & 27,611 & 5860 & 5862 \\
        \bottomrule
    \end{tabular}
    }
    \label{tab:dataset_stats}
\end{table}

\subsubsection{Tokenizer and model training:} We evaluate the effect of different tokenizers and their vocabularies on the quality of report summarisation and memory requirements to train a model.
\
We evaluate 3 types of BPE tokenizers, all trained independently prior to any LM training.
\textbf{General:} First, we adopt the GPT-2 tokenizer~\cite{radford_gpt}, trained on varied texts by OpenAI, which has a general-purpose vocabulary of size $V\approx50$k. \textbf{Medical:} We train a tokenizer on varied medical corpus of PubMed abstracts (similarly to \cite{Gu_2022, Boecking_2022_ms}) resulting in vocabulary of size $V=30k$. \textbf{Specific:} Finally, we train 3 domain-specific tokenizers, one per dataset, which results in vocabulary sizes $V$ ranging from 9 to 11k tokens.
Values of $V$ for the domain-specific tokenizers are determined dynamically during training by including only those words (or subwords) that appear at least three times in the training corpus.

Each experiment begins by using a trained tokenizer to preprocess the data. An autoregressive LM is subsequently trained to generate the Conclusion from the Findings tokens (Fig.~\ref{fig:model}).
The LM hyper-parameters used are: $N=8$, $H =8$, $D =512$, $D_{ff} =2048$, where $D_{ff}$ is the dimensionality of feed forward layers. 
\
In each experiment, we set maximum sequence length $S$ for the Transformer's input at the 90th (99th for MIMIC) percentile of the report length. Tokenizers that fragment words more frequently require larger $S$.
Given a limited memory budget (48 GBs of 1 NVIDIA A6000), as the tokenizer affects how much memory is required per sample, we adjusted the batch sizes in each experiment to maximize memory utilisation. 
Details on sequence length and batch size used are in Supplementary Table 1.
In addition to training the LM  \textbf{from scratch} on a dataset, we also investigate the effect of tokenization when the LM is \textbf{pre-trained} on a biomedical corpus (PubMed abstracts) before it is \emph{fine-tuned} on the target dataset. Such pre-training is commonly used to enhance LM performance when training data for the target task is limited~\cite{Liu_review_2023}.

\subsubsection{Metrics:} We assess overall language quality using common NLP metrics: BLEU (BL-n)~\cite{bleu}, METEOR (M)~\cite{meteor} and \mbox{ROUGE-L} (R-L)~\cite{rouge_2004}. 
To assess clinical efficacy (CE), we extract disease classification labels from the generated and reference conclusions using automated labellers
\cite{chexbert_2020, Hamamci_ctclip_2024}, 
and compare them to compute F1 scores.
We calculate micro and macro averages of F1 over 14 classes in MIMIC-CXR and 18 classes in CT-RATE, using the CheXbert~\cite{chexbert_2020} and RadBERT labellers~\cite{Hamamci_ctclip_2024} respectively. No text-based disease labeller exists for PET-CT reports, thus this metric is unavailable in this dataset. 
As measure of general factual consistency, we calculate RadGraph-XL F1~\cite{radgraphXL_2024} over all entity and relation types.
We also report cosine similarity using a biomedical BERT~\cite{Boecking_2022_ms}.
All metrics range from 0 to 1, and higher values indicate a closer match to the reference. 

\subsection{Results} 

\begin{table}[t]
\centering
\renewcommand{\arraystretch}{0.9}
\caption{Performance using different vocabularies when LMs are trained from scratch. Best results per dataset in bold.
}
\resizebox{\textwidth}{!}{
\begin{tabular}{
    @{}>{\raggedright}m{0.12\textwidth}
    *{1}{>{\centering\arraybackslash}m{0.12\textwidth}}
    *{1}{>{\centering\arraybackslash}m{0.11\textwidth}}
    *{1}{>{\centering\arraybackslash}m{0.13\textwidth}}
    *{3}{>{\centering\arraybackslash}m{0.12\textwidth}}
    *{4}{>{\centering\arraybackslash}m{0.15\textwidth}}
    @{}
}
\toprule
\multirow{3}{0.15\textwidth}{\textbf{Eval. Dataset}} & \multirow{3}{0.12\textwidth}{\centering\textbf{Tokenizer}} & \multirow{3}{0.14\textwidth}{\centering\textbf{\mbox{Pretrain}}} & \multirow{3}{0.12\textwidth}{\centering\textbf{Vocab size}} &
\multicolumn{3}{c}{\textbf{NLP Metrics}} & 
\multicolumn{4}{c}{\textbf{CE Metrics}} \\ 
\cmidrule(rl){5-7} \cmidrule(l){8-11}
 &  &  &   & 
\textbf{BL-1} & \textbf{M} & \textbf{R-L} & 
\textbf{F1 \textit{macro}} & \textbf{F1 \textit{micro}} & 
\textbf{RadXL-F1} & \textbf{Cos. Sim.} \\
\midrule
\midrule
\multirow{3}{*}{MIMIC} 
& Specific & \xmark & 9.1k & \textbf{0.273} & \textbf{0.364} & \textbf{0.344} & \textbf{0.514} & \textbf{0.604} & \textbf{0.270} & \textbf{0.813} \\
& Medical & \xmark & 30k & 0.262 & 0.353 & 0.325 & 0.480 & 0.594 & 0.243 & 0.809 \\
& General & \xmark & 50k & 0.162 & 0.262 & 0.214 & 0.399 & 0.504 & 0.164 & 0.700 \\
\midrule
\multirow{3}{*}{CT-RATE}
& Specific & \xmark & 9.4k & \textbf{0.369} & \textbf{0.450} & \textbf{0.423} & \textbf{0.477} & \textbf{0.547} & \textbf{0.196} & \textbf{0.835} \\
& Medical & \xmark & 30k & 0.344 & 0.423 & 0.400 & 0.458 & 0.544 & 0.175 & 0.819 \\
& General & \xmark & 50k & 0.339 & 0.415 & 0.395 & 0.402 & 0.479 & 0.192 & 0.822 \\
\midrule
\multirow{3}{*}{PET-CT}
& Specific & \xmark & 11.0k & \textbf{0.241} & \textbf{0.304} & \textbf{0.298} & - & - & \textbf{0.126} & \textbf{0.844} \\
& Medical & \xmark & 30k & 0.234 & 0.294 & 0.283 & - & - & 0.116 & 0.838 \\
& General & \xmark & 50k & 0.176 & 0.241 & 0.234 & - & - & 0.064 & 0.793 \\
\bottomrule
\end{tabular}
}
\label{tbl:metrics_bpe}
\end{table}

We evaluate the impact of different tokenizers on report summarisation performance using held-out test sets from MIMIC-CXR, CT-RATE, and PET-CT. Table~\ref{tbl:metrics_bpe} shows results for LMs trained from scratch on each dataset with different tokenizers, 
providing several insights. First, both tokenizers trained on medical text (generic \emph{Medical} and domain-\emph{Specific}) consistently outperform general-purpose tokenizers across all metrics when no pre-training is used (Table \ref{tbl:metrics_bpe}). The domain-\emph{Specific} outperforms the \emph{Medical} trained on PubMed. This is consistent across reports of all imaging types.
Results suggest that when training an LM for a specific task from scratch, practitioners should create domain-specific tokenizers instead of adopting general-purpose vocabularies.

\begin{table}[b]
\centering
\renewcommand{\arraystretch}{1.2}
\caption{Results for LMs pre-trained on PubMed followed by dataset-specific fine-tuning. Best results per dataset in bold. Relative (\%) change from models without pre-training shown in brackets.}
\resizebox{\textwidth}{!}{
\begin{tabular}{
    @{}>{\raggedright}m{0.12\textwidth}
    *{1}{>{\centering\arraybackslash}m{0.12\textwidth}}
    *{1}{>{\centering\arraybackslash}m{0.1\textwidth}}
    *{1}{>{\centering\arraybackslash}m{0.11\textwidth}}
    *{3}{>{\raggedright\arraybackslash}m{0.16\textwidth}}
    *{4}{>{\raggedright\arraybackslash}m{0.16\textwidth}}
    @{}
}
\toprule
\multirow{3}{0.15\textwidth}{\textbf{Eval. Dataset}} & \multirow{3}{0.12\textwidth}{\centering\textbf{Tokenizer}} & \multirow{3}{0.13\textwidth}{\centering\textbf{Pretrain}} & \multirow{3}{0.12\textwidth}{\centering\textbf{Vocab size}} &
\multicolumn{3}{c}{\textbf{NLP Metrics ($\Delta$\%)}} & 
\multicolumn{4}{c}{\textbf{CE Metrics ($\Delta$\%)}} \\ 
\cmidrule(rl){5-7} \cmidrule(l){8-11}
 &  &  &   & 
\textbf{BL-1} & \textbf{M} & \textbf{R-L} & 
\textbf{F1 \textit{macro}} & \textbf{F1 \textit{micro}} & 
\textbf{RadXL-F1} & \textbf{Cos. Sim.} \\
\midrule
\midrule
\multirow{3}{*}{MIMIC} 
& Specific & \cmark & 9.1k & \makebox{\textbf{0.285} {\scriptsize(+4.4)}} & \makebox{\textbf{0.380} {\scriptsize(+4.4)}} & \makebox{\textbf{0.359} {\scriptsize(+4.4)}} & \makebox{\textbf{0.521} {\scriptsize(+1.4)}} & \makebox{\textbf{0.622} {\scriptsize(+3.0)}} & \makebox{\textbf{0.283} {\scriptsize(+4.8)}} & \makebox{\textbf{0.817} {\scriptsize(+0.5)}} \\
& Medical & \cmark & 30k & \makebox{0.284 {\scriptsize(+8.4)}} & \makebox{\textbf{0.380} {\scriptsize(+7.6)}} & \makebox{0.355 {\scriptsize(+9.2)}} & \makebox{0.518 {\scriptsize(+7.9)}} & \makebox{0.621 {\scriptsize(+4.5)}} & \makebox{0.268 {\scriptsize(+10.3)}} & \makebox{0.814 {\scriptsize(+0.6)}} \\
& General & \cmark & 50k & \makebox{0.167 {\scriptsize(+3.1)}} & \makebox{0.261 {\scriptsize(–0.4)}} & \makebox{0.220 {\scriptsize(+2.8)}} & \makebox{0.417 {\scriptsize(+4.5)}} & \makebox{0.527 {\scriptsize(+4.6)}} & \makebox{0.169 {\scriptsize(+3.0)}} & \makebox{0.721 {\scriptsize(+3.0)}} \\
\midrule
\multirow{3}{*}{CT-RATE} 
& Specific & \cmark & 9.4k & \makebox{\textbf{0.403} {\scriptsize(+9.2)}} & \makebox{\textbf{0.492} {\scriptsize(+9.3)}} & \makebox{\textbf{0.468} {\scriptsize(+10.6)}} & \makebox{0.564 {\scriptsize(+18.2)}} & \makebox{0.616 {\scriptsize(+12.6)}} & \makebox{0.229 {\scriptsize(+16.8)}} & \makebox{\textbf{0.844} {\scriptsize(+1.1)}} \\
& Medical & \cmark & 30k & \makebox{0.376 {\scriptsize(+9.3)}} & \makebox{0.467 {\scriptsize(+10.4)}} & \makebox{0.435 {\scriptsize(+8.8)}} & \makebox{0.540 {\scriptsize(+17.9)}} & \makebox{0.598 {\scriptsize(+9.9)}} & \makebox{0.191 {\scriptsize(+9.1)}} & \makebox{0.830 {\scriptsize(+1.3)}} \\
& General & \cmark & 50k & \makebox{\textbf{0.403} {\scriptsize(+18.9)}} & \makebox{\textbf{0.492} {\scriptsize(+18.6)}} & \makebox{0.451 {\scriptsize(+14.2)}} & \makebox{\textbf{0.568} {\scriptsize(+41.3)}} & \makebox{\textbf{0.620} {\scriptsize(+29.4)}} & \makebox{\textbf{0.243} {\scriptsize(+26.6)}} & \makebox{0.840 {\scriptsize(+2.2)}} \\
\midrule
\multirow{3}{*}{PET-CT} 
& Specific & \cmark & 11.0k & \makebox{\textbf{0.238} {\scriptsize(–1.2)}} & \makebox{\textbf{0.302} {\scriptsize(–0.7)}} & \makebox{\textbf{0.280} {\scriptsize(–6.0)}} & - & - & \makebox{\textbf{0.127} {\scriptsize(+0.8)}} & \makebox{\textbf{0.842} {\scriptsize(–0.2)}} \\
& Medical & \cmark & 30k & \makebox{\textbf{0.238} {\scriptsize(+1.7)}} & \makebox{0.300 {\scriptsize(+2.0)}} & \makebox{0.277 {\scriptsize(–2.1)}} & - & - & \makebox{0.120 {\scriptsize(+3.4)}} & \makebox{0.840 {\scriptsize(+0.2)}} \\
& General & \cmark & 50k & \makebox{\textbf{0.238} {\scriptsize(+35.2)}} & \makebox{0.299 {\scriptsize(+24.1)}} & \makebox{0.261 {\scriptsize(+11.5)}} & - & - & \makebox{0.123 {\scriptsize(+92.2)}} & \makebox{0.833 {\scriptsize(+5.0)}} \\
\bottomrule
\end{tabular}
}
\label{tbl:metrics_bpe_pp_percent_scriptsize}
\end{table}

When pre-training a LM before fine-tuning it for a task of interest, a natural question arises: what tokenizer should one use? 
To assess the impact of tokenizer choice in this scenario, we pre-train LMs using each of the 3 tokenizers on PubMed abstracts, followed by fine-tuning and evaluation on the target dataset. 
Results are shown in Table \ref{tbl:metrics_bpe_pp_percent_scriptsize}.
Pre-training generally improves performance in comparison to training from scratch, with the largest gains observed for LMs using the \emph{General} or \emph{Medical} tokenizers. In some cases, these models approach the performance of LMs that use the domain-\emph{Specific} tokenizer. 
We conjecture that this is because more general tokenizers often break domain-specific terms into many subword units, and pre-training helps the LM learn meaningful relationships between these fragments. 
In contrast, domain-specific tokenizers preserve such terms as whole units, avoiding this issue.
LMs using domain-specific tokenizers retain high overall performance across most settings. 
This suggests that when the target task is known, it is still recommended to design the tokenizer accordingly before pre-training, as it will shape the model's representations throughout.
\begin{figure}[t]
\centering
\includegraphics[width=\textwidth]{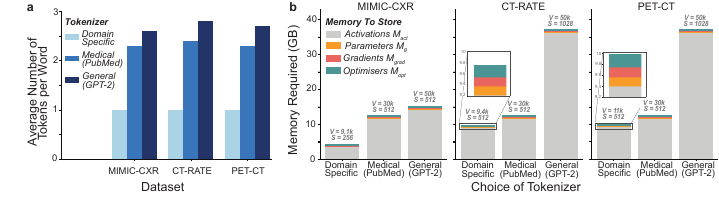}
\caption{
(a) Average number of tokens per word for each dataset and tokenizer.
(b) Memory required to train LMs for the different tokenizers, calculated using Eq.~\ref{eq:mem_tot} for const. batch size B=32 (max possible for \emph{General} in our GPU).
Domain-specific tokenizers produce fewer subword splits, allowing shorter sequences ($S$) which combined with smaller vocabulary size $V$ results in lower memory usage. 
}
\label{fig:mem}
\end{figure}

\begin{table}[b]
    \centering
    \renewcommand{\arraystretch}{0.8}
    \caption{Examples of how 3 tokenizers fragment medical words to tokens. 
    }
    \resizebox{\textwidth}{!}{
    \begin{tabular}{p{4cm} p{4cm} p{4cm}}
        \toprule
        \textbf{Domain specific} & \textbf{Biomedical} & \textbf{General} \\
        \midrule
        bronchovasculature & broncho-vasculature   &  b-ron-ch-ov-as-cul-ature\\
        multinodular & multin-odular & mult-in-od-ular \\
        sternomastoid  & sterno-mastoid  & st-ern-om-ast-oid  \\
        adenopathy  & adeno-pathy  & aden-opathy  \\
        consolidation & consolidation & cons-olid-ation \\
        \bottomrule
    \end{tabular}}
    \label{tab:words}
\end{table}

Beyond model performance, domain-specific tokenization also reduces the memory required to train the LM. 
As shown in Fig.~\ref{fig:mem}a and Table~\ref{tab:words}, the domain-specific tokenizer breaks medical terms into fewer, more meaningful tokens - capturing all words that appear at least 3 times in the dataset, compared to an average of 2.7 tokens per word with the general tokenizer. 
This more efficient tokenization allows for shorter sequence lengths $S$ which, together with a smaller vocabulary $V$, means that domain-specific tokenizers consistently require less memory than a general-purpose tokenizer, as shown in Fig~\ref{fig:mem}b. We calculate memory requirements using Eq~\ref{eq:mem_tot}. 
Domain-specific tokenisation provides efficiency gains which do not compromise the quality of generated reports (Table~\ref{tbl:metrics_bpe} and~\ref{tbl:metrics_bpe_pp_percent_scriptsize}).

\section{Conclusion}

This study analysed the impact of tokenizers on language modelling for summarization of radiology reports of X-rays, CTs and, oncology PET-CTs. Results show that domain-specific tokenizers with vocabularies tailored to the radiology task of interest improve LMs performance in comparison to more general tokenizers. This effect is especially strong when LM models are trained from scratch.
When we pre-train the LMs, the performance gap is reduced, though domain-specific tokenizers still retain consistently high performance across tasks.
The study also provided a theoretical analysis of how tokenizers affect memory requirements of LMs, showing that domain-specific tokenizers require substantially less memory than general natural language or medical tokenizers. This can enable LM training with more accessible GPU hardware.
These findings provide valuable insights for machine learning practitioners, underscoring it can be advantageous to create tokenizers for the specific task of interest, rather than adopting generic ones, prior to training radiological language models.

\section{Acknowledgements}
Hermione Warr and Yasin Ibrahim are supported by the EPSRC Centre for Doctoral Training in Health Data Science (EP/S02428X/1).
This research has been conducted using NHS data accessed via the Thames Valley and Surrey Secure Data Environment (TVS SDE), as part of the NHS Research Secure Data Environment Network. The NHS Research SDE Network is jointly funded by NHS England, the Department of Health and Social Care (DHSC) and the Department for Business, Energy and Industrial Strategy (BEIS). Data collection and processing within TVS SDE is also funded in part by the NIHR Oxford Biomedical Research Centre at Oxford University Hospitals NHS FT.
%
%
%
\bibliographystyle{splncs04}
\bibliography{Scribe}

\end{document}